\documentclass[letterpaper]{article} 
\usepackage[preprint]{arxiv2026}  
\usepackage[hyphens]{url}  
\usepackage{graphicx} 
\usepackage{natbib}  
\usepackage{caption} 
\usepackage{booktabs}

\usepackage{amsmath,amssymb}
\usepackage{xspace}
\usepackage[table]{xcolor} 

\makeatletter
\renewcommand{\@seccntformat}[1]{%
  \csname the#1\endcsname.\quad}
\makeatother

\newcommand{\method}{SCOPE\xspace}
\newcommand{\ctrate}{CT-RATE\xspace}
\newcommand{\radchest}{RadChestCT\xspace}

\title{Semantically Calibrated Evidence Composition for CT Vision-Language Learning}

\author{
Guoliang You\textsuperscript{1},
Haifan Gong\textsuperscript{2}\corresponding,
Xiaomeng Chu\textsuperscript{3}\corresponding
}

\affiliations{
\textsuperscript{1}University of Pennsylvania,
\textsuperscript{2}Harvard Medical School
\textsuperscript{3}Yale University
}

\begin{document}
\maketitle

\begin{abstract}
Learning transferable representations from CT-report pairs requires combining whole-volume context with anatomy-specific evidence. Existing methods typically emphasize either global CT-report alignment or fine-grained anatomy-level correspondence. Global alignment preserves broad study context but leaves the contribution of localized evidence implicit, whereas anatomy-level alignment explicitly grounds local findings but does not specify how independently represented evidence should interact, acquire study-level meaning, and contribute to a global CT representation. To address this gap, we propose \method (Semantic Calibration Of comPosed Evidence), a framework for semantically calibrated evidence composition in CT vision-language learning. Under organ-specific report supervision, mask-guided queries with fixed anatomical identities extract context-aware organ evidence from shared, uncropped volumetric features, while an unrestricted global query retains access to whole-volume context. The global query then drives Local-Global Coupling to compose the organ evidence into a unified evidence representation. The composed evidence is subsequently calibrated using the diagnostic summary, providing study-level semantic supervision beyond local organ descriptions, and is finally integrated as a controlled residual into a context-preserving whole-volume representation aligned with the complete report. This progressive pathway connects localized evidence with study-level semantics without reducing the CT representation to a predefined set of organs. 
On CT-RATE and RadChestCT, \method achieves macro AUCs of 85.0 and 72.2, respectively, outperforming the previous SOTA by 7.2 and 4.2, while also yielding substantial gains in linear probing and cross-modal retrieval. These results demonstrate the effectiveness of semantically calibrated evidence composition.
\end{abstract}

\section{Introduction}
\begin{figure}[!t]
    \centering
    \includegraphics[width=\columnwidth]{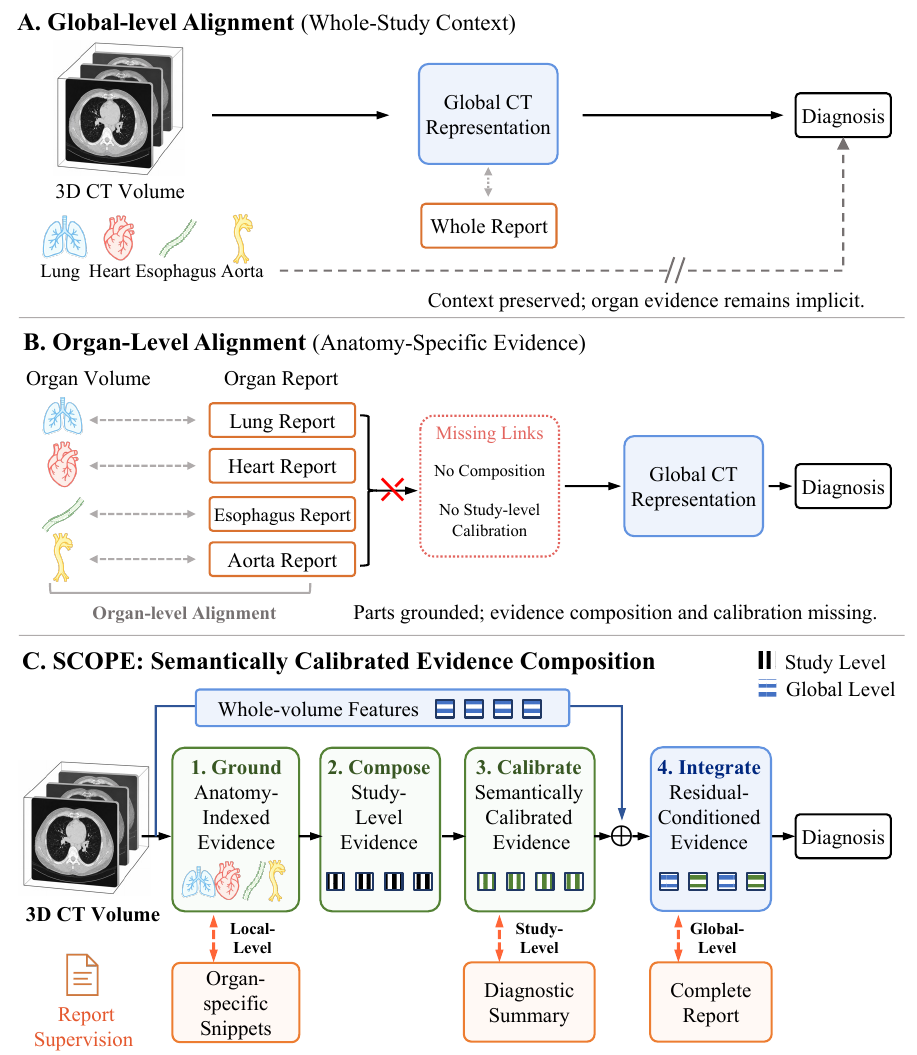}
    \caption{Global-level alignment preserves whole-study context while leaving anatomy-specific evidence implicit (A). Organ-level alignment grounds localized evidence but does not explicitly compose it under whole-volume context or calibrate it toward study-level semantics (B). \method addresses both gaps by grounding anatomy-indexed evidence, composing it through Local--Global Coupling (LGC), calibrating it with the diagnostic summary, and integrating it as a controlled residual into the whole-volume base (C).}
    \label{fig:motivation}
\end{figure}

Computed tomography (CT) plays a central role in the assessment of thoracic diseases, yet expert annotation of volumetric scans is costly and difficult to scale~\citep{langlotz2024merlin,gong2024intensity,huang2025bcnet,wang2026costal,chen2023cancerunit}. Radiology reports are routinely paired with these scans and record both organ-specific observations and study-level interpretations, making them a natural source of supervision for transferable CT representations~\cite{hamamci2024generalist,shui2025fvlm}. Effective CT vision-language learning should therefore preserve unrestricted whole-volume context while retaining the organ-specific evidence that supports study-level semantics.

Global CT-report alignment addresses the whole-study side of this requirement by aligning whole-volume and report-level representations~\cite{hamamci2024generalist,langlotz2024merlin,wald2025colipri}. Because the report semantically organizes findings across multiple anatomical regions, this objective is effective for learning broad study-level correspondence while preserving unrestricted whole-volume context. Its limitation is not that organ information is necessarily absent, but that the objective does not require anatomy-specific evidence to be explicitly represented or to contribute through an identifiable pathway to the embedding used for prediction.

Fine-grained alignment addresses the complementary side by grounding representations in localized evidence~\cite{huang2021gloria,muller2022joint}; fVLM extends this idea to explicit anatomy-level targets in 3D CT~\cite{shui2025fvlm}. Yet two mismatches remain. Structurally, independently aligned organ representations are not automatically composed into coherent study-level evidence, and their direct aggregation may fail to retain cross-organ relations, broader spatial context, or findings outside the predefined anatomy set. Semantically, their local semantics are not automatically compatible with study-level interpretation. As shown in Figure~\ref{fig:motivation}, the missing mechanism is therefore an explicit \emph{evidence-composition and semantic-calibration interface} that composes anatomy-indexed evidence under whole-volume context and aligns it with study-level interpretation.

Radiology reports provide a natural supervisory hierarchy for this transition. Organ-specific snippets describe localized findings, the diagnostic summary expresses their joint study-level meaning, and the complete report preserves broader findings and examination context. Together, these views provide progressively broader semantic targets, from anatomy-specific observations to whole-examination interpretation.

We therefore introduce \method (\emph{Semantic Calibration Of comPosed Evidence}), which instantiates this transition through \emph{Ground--Compose--Calibrate--Integrate}. It first \emph{grounds} learnable queries with fixed organ identities in shared, uncropped 3D visual features using mask-guided refinement and organ-specific report supervision. Alongside the organ queries, an unrestricted global query retains access to whole-volume context and drives Local--Global Coupling (LGC) to \emph{compose} anatomy-indexed evidence through attention rather than independent pooling or summation, addressing the structural mismatch. During training, diagnostic-summary contrastive supervision \emph{calibrates} the composed evidence toward study-level semantics, addressing the semantic mismatch. Finally, the calibrated evidence is projected and \emph{integrated} as a controlled residual into a direct whole-volume base, and the resulting residual-conditioned representation is aligned with the complete report.

The direct whole-volume base and the evidence residual play complementary roles. The former preserves unrestricted examination context, including findings beyond the predefined organs, cross-organ relations, and background information, while the latter provides an explicit route for semantically calibrated composed evidence to contribute to the final representation. 

In summary, our main contributions are as follows:
\begin{itemize}
    \item We identify an \emph{evidence-composition and semantic-calibration gap} in CT vision-language learning: organ-level objectives ground anatomy-indexed representations, but do not automatically compose them into coherent study-level evidence or align their local semantics with study-level interpretation.
    \item We introduce \method, a report-mediated \emph{Ground-Compose-Calibrate-Integrate} framework that grounds anatomy-specific evidence with mask-guided queries, composes it with global context, calibrates it toward study-level semantics via diagnostic-summary supervision, and integrates it into a context-preserving whole-volume representation.
    \item We achieve 85.0 and 72.2 AUC on CT-RATE and \radchest, respectively, together with complementary retrieval, organ-recognition, and organ-occlusion analyses that assess cross-modal structure and anatomy-indexed representation learning.
\end{itemize}

\section{Related Work}

\begin{figure*}[!t]
    \centering
    \includegraphics[width=0.95\textwidth]{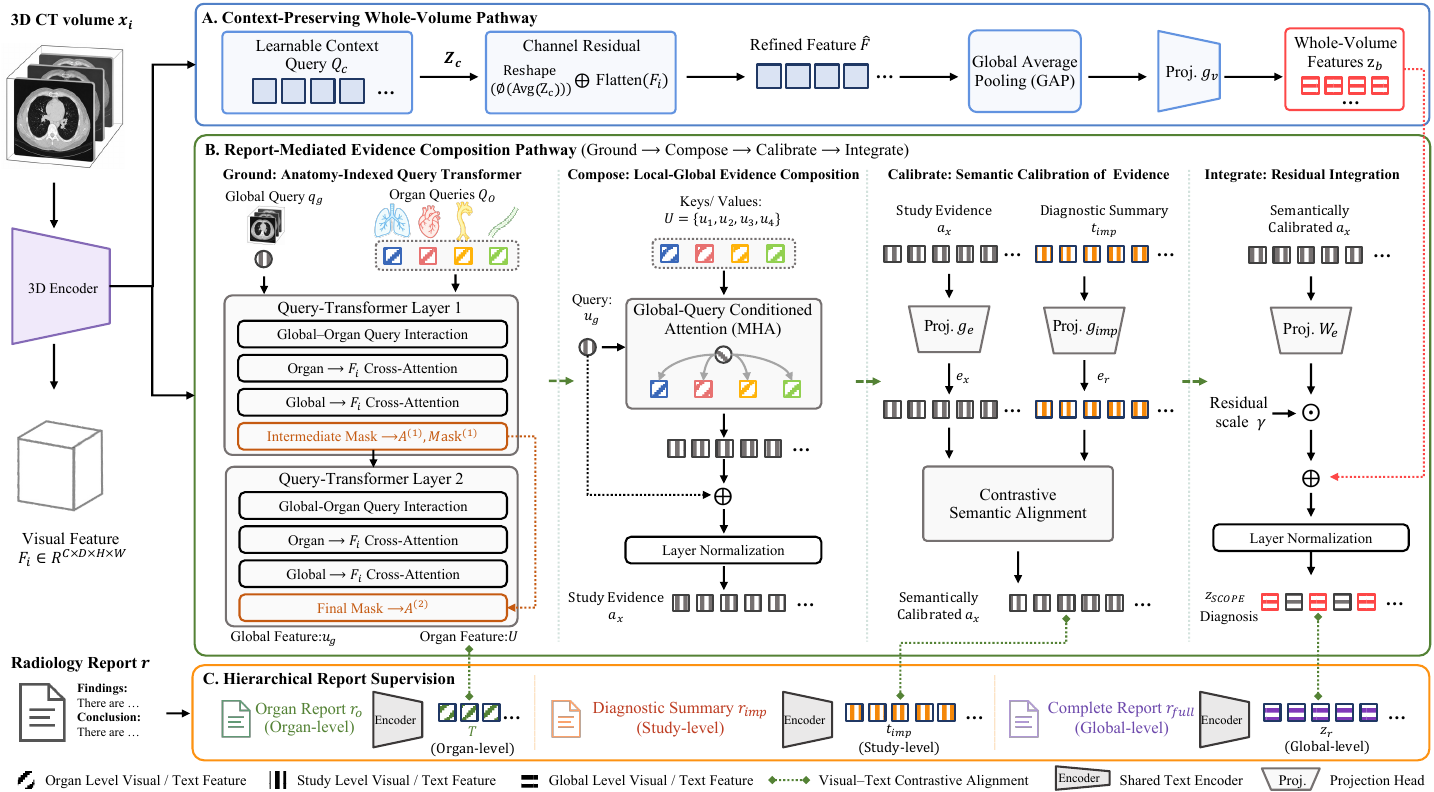}
    \caption{Architecture of \method. The 3D image encoder produces visual features $F$; context refinement yields $\hat F$ and the direct whole-volume representation $z_b$. In parallel, the anatomy-indexed query transformer reads uncropped $F$ and returns context-aware organ tokens $U$ and an unrestricted global query $u_g$. Image Local-Global Evidence Composition (LGC) uses $Q=u_g$ and $K,V=U$ to address the structural composition mismatch by forming study-level evidence $a_x$. Diagnostic-summary supervision addresses the semantic transition mismatch by calibrating $a_x$ during training, after which it enters $z_{\mathrm{\method}}=\mathrm{LN}(z_b+\gamma W_ea_x)$ as a controlled residual. Organ snippets ground $U$, and the complete report aligns the final representation. Supporting objectives are separated in the final strip. The residual-conditioned representation is used for diagnosis, retrieval, and external evaluation.}
    \label{fig:architecture}
\end{figure*}

\subsection{Medical Vision-Language Learning}

Vision-language pre-training learns transferable medical representations from naturally paired images and clinical text~\cite{zhang2025multimodal,liu2025visual}. Early methods extended contrastive image-text learning to radiology~\cite{dong2022survey,zhang2022convirt,tiu2022chexzero,wang2022medclip,zhou2022refers}, while subsequent approaches exploited global-local correspondence, medical knowledge, masked modeling, and prompt-based learning to strengthen cross-modal representations~\cite{huang2021gloria,boecking2022cxrbert,chen2022arl,chen2022m3ae,wu2023medklip,chen2023ptunifier,gong2021crossmodal,gong2022vqamix}. 
The paradigm has recently expanded from 2D radiographs to volumetric imaging. T3D~\cite{liu2023t3d} exploits multi-view consistency for 3D medical vision-language pre-training, while CT-CLIP~\cite{hamamci2024generalist} and Merlin~\cite{langlotz2024merlin} learn transferable CT representations from large-scale volumetric image-text data. More recent foundation models further investigate scalable and generalizable representation learning for CT and broader radiology applications~\cite{ct-fm,codella2024medimageinsight,dancette2025curia,wald2025colipri}.

However, existing medical VLP methods primarily improve global image-text correspondence or general-purpose multimodal representations. In contrast, \method focuses on how explicit localized evidence should contribute to a global CT representation: it composes anatomy-indexed evidence under whole-volume context, semantically calibrates the composed evidence using the diagnostic summary, and integrates it into the global representation rather than replacing global CT-report alignment.

\subsection{Fine-Grained CT Representation Learning}

Fine-grained vision-language learning aims to expose clinically meaningful local or organ-level evidence~\cite{huang2025bcnet,gong2025domain,gong2025boundary,wang2026costal} beyond global image-report correspondence~\cite{boecking2022cxrbert,zhou2022refers,wu2023medklip,chen2022m3ae}. GLoRIA~\cite{huang2021gloria} models global-local relationships between image regions and report words, while localized representation learning~\cite{muller2022joint} and knowledge-enhanced approaches~\cite{chen2022arl,wu2023medklip} further strengthen correspondence between visual findings and clinical semantics. In volumetric CT, fVLM~\cite{shui2025fvlm} explicitly aligns anatomy-specific visual representations with corresponding organ-level report descriptions, providing direct anatomical supervision for CT representation learning.

However, prior fine-grained and anatomy-aware methods primarily focus on \emph{grounding} or improving the representation of localized anatomical information. \method instead treats grounded organ evidence as an intermediate representation and addresses what happens afterward. Local-Global Coupling explicitly composes evidence across anatomical regions under global context, diagnostic-summary supervision calibrates the composed representation toward study-level semantics, and residual integration incorporates it into an unrestricted whole-volume representation. Thus, the central contribution of \method is evidence composition and semantic calibration rather than anatomical grounding itself.

\section{Method}
\label{sec:method}

\subsection{Problem Formulation and Overview}
Let $x$ denote a 3D CT volume and $r$ its paired report. From the structured CT-RATE records, we construct three nested textual views at increasing semantic scales. Organ-specific snippets $\{r_o\}_{o\in\mathcal O}$, where $\mathcal O=\{\mathrm{lung},\mathrm{heart},\mathrm{esophagus},\mathrm{aorta}\}$, aggregate available organ-level text from \texttt{Findings} and \texttt{Conclusion}. The study-level diagnostic summary $r_{\mathrm{imp}}$ uses the overall \texttt{Conclusion}, with organ conclusions as fallback, while the complete-report view $r_{\mathrm{full}}$ combines all available global \texttt{Findings} and \texttt{Conclusion} fields. Together, these views provide hierarchical supervision from anatomy-specific findings, to study-level diagnostic semantics, to comprehensive examination context.

\method exploits this hierarchy through a \emph{Ground-Compose-Calibrate-Integrate} pipeline in Figure~\ref{fig:architecture}. A context-preserving whole-volume pathway first produces the global-level whole-volume features $z_b$, while anatomy-indexed queries generate organ tokens $U$ and an unrestricted global query $u_g$ retains access to the complete visual memory. Local-Global Coupling (LGC) then uses $u_g$ as the query and $U$ as keys and values to \emph{compose} localized evidence under whole-volume context, yielding $a_x$. Diagnostic-summary contrastive supervision subsequently \emph{calibrates} $a_x$ toward study-level semantics, after which the projected evidence residual $\gamma W_ea_x$ is \emph{integrated} into $z_b$ to obtain the residual-conditioned representation $z_{\mathrm{SCOPE}}$. In this way, \method serves as a report-mediated evidence composition and calibration interface rather than a generic fusion of independently learned global and local representations; the individual grounding, LGC, calibration, and residual-integration components are detailed below.

\subsection{Context-Preserving Whole-Volume Pathway}

A R3D-18 backbone~\cite{hara2018r3d} maps $x$ to
$F\in\mathbb{R}^{B\times C\times D\times H\times W}$, which is flattened as
$F_{\mathrm{flat}}=M\in\mathbb{R}^{B\times N\times C}$, where $N=DHW$ is the number of spatial positions. A bank of $K$ learnable context-refinement tokens $Q_c$ reads $M$, and their mean decoded context is projected and broadcast as a channel residual:
\begin{equation}
\begin{aligned}
Z_c &= \mathrm{Decoder}(Q_c,M),\\
\hat F &= F+\alpha\,\mathrm{reshape}\!\left(
\phi\!\left(\operatorname{Avg}_{K}(Z_c)\right)
\right).
\end{aligned}
\label{eq:context_refinement}
\end{equation}
Here, $\operatorname{Avg}_{K}$ averages over the $K$ decoded context tokens, $\phi$ is an MLP, and $\alpha$ is a learnable residual scale. Equation~\ref{eq:context_refinement} performs sample-wise channel refinement rather than position-specific spatial refinement. The direct pathway reads the refined feature map $\hat F$, whereas the anatomy-indexed query transformer reads the original encoder feature map $F$; both originate from the same anatomy-supervised image encoder.

Let $E_t$ be the shared text encoder, $g_v$ and $g_t$ the visual and text projection heads, and $\mathrm{GAP}$ global average pooling. We obtain the whole-volume visual representation $z_b$ and its complete-report counterpart $z_r$ as:
\begin{equation}
z_b=g_v(\mathrm{GAP}(\hat F)),\qquad
z_r=g_t(E_t(r_{\mathrm{full}})).
\label{eq:global_base}
\end{equation}
Since $z_b$ is directly derived from the full encoded volume, it preserves unrestricted examination context, including findings outside the four predefined organs and relationships across anatomical regions.

\subsection{Ground: Anatomy-Indexed Query Transformer}

The query transformer introduces a bank of learnable organ queries
$Q_{\mathcal O}=\{q_o\}_{o\in\mathcal O}$ with fixed anatomical assignments, together with an unrestricted global query $q_g$. We initialize $U^{(0)}=Q_{\mathcal O}$ and $u_g^{(0)}=q_g$. Each organ output is aligned with its corresponding report snippet, while mask-guided refinement encourages the query to remain associated with its assigned anatomy.

Within each block, the organ queries exchange information, the global query attends to the organ set, and the updated global context is propagated back to the organ queries. Both query types additionally cross-attend to the shared, uncropped visual features $F$. This interaction yields context-aware anatomy-indexed tokens together with an unrestricted global coordinator. The final global query therefore serves not as an auxiliary classification token, but as the global context used by LGC to compose localized evidence.
Let $(U^{(\ell)},u_g^{(\ell)})$ denote the outputs of query-transformer layer $\ell$. After each layer, organ-memory similarity is computed as
\begin{equation}
A^{(\ell)}=\frac{U^{(\ell)}F_{\mathrm{flat}}^\top}{\sqrt{C}},
\qquad \ell=1,2.
\label{eq:aaq_routing}
\end{equation}
We take $U=U^{(2)}=\{u_o\}_{o\in O}$ and $u_g=u_g^{(2)}$ as the final organ-token set and global-query output, respectively. Layer~1 performs unmasked organ-to-memory cross-attention, whose sigmoid-thresholded logits $A^{(1)}$ define the routing mask for Layer~2 during both training and inference. Layer~2 restricts only organ-to-memory attention with this mask, while the global query remains fully connected to the visual memory. The final logits $A^{(2)}$ are supervised by the organ masks through the mask-grounding loss.

This design differs from crop-first modeling~\cite{shui2025fvlm}. Because the full volume is encoded before mask-guided attention, first-layer organ queries initially access the complete visual memory, while the global query remains unrestricted throughout both layers. The predicted masks guide anatomical identity and attention rather than defining cropped encoder inputs, and organ-specific report snippets provide the corresponding semantic supervision. Consequently, $U$ remains anatomy-indexed yet context-aware, while $u_g$ supplies whole-volume context for subsequent evidence composition.

Additionally, the organ-specific snippets are encoded as $t_o=E_t(r_o)$, forming the organ-level text-token set $T=\{t_o\}_{o\in\mathcal O}$. Let $\mathcal O_v\subseteq\mathcal O$ denote the organs with sufficient valid samples in the batch, and let $\tau_{\mathrm{loc}}$ denote the shared local temperature.
For organ-level semantic grounding, let $\bar U_o$ and $\bar T_o$ collect the normalized projected features of the visual token $u_o$ and text token $t_o$, respectively, over the valid samples for organ $o$.
We apply an organ-wise bidirectional contrastive objective:
\begin{equation}
\mathcal L_{\mathrm{local}}
\!=\!\!
\sum_{o\in\mathcal O_{\mathrm v}}
\frac{1}{2}
\left[
\mathrm{CE}\!\left(
\bar U_o\bar T_o^\top/\tau_{\mathrm{l}},P_o
\right)
\!+\!
\mathrm{CE}\!\left(
\bar T_o\bar U_o^\top/\tau_{\mathrm{l}},P_o^\top
\right)
\right].
\label{eq:local_itc}
\end{equation}
Here $P_o$ is constructed from CT-RATE labels to augment paired positives with semantically matched organ-level samples, while unmatched pairs serve as negatives. This label-aware target reduces false negatives among cases with similar findings and provides a richer supervision signal than one-to-one pairing alone.

\subsection{Compose: Local-Global Evidence Composition}

Image-side Local-Global Coupling performs global-query-conditioned evidence composition. The global query output $u_g$ serves as the attention query, while the anatomy-indexed token set $U$ provides the keys and values:
\begin{equation}
a_x=\mathrm{LGC}_v(u_g,U)
=\mathrm{LN}\!\left(u_g+\mathrm{MHA}(u_g,U,U)\right).
\end{equation}
The attention term conditions the global representation on anatomy-indexed evidence, while the identity residual preserves information that $u_g$ has obtained directly from the full visual memory. Thus, $a_x$ is a global-context-conditioned composition of localized organ evidence rather than a purely local or exclusively organ-derived representation.

This construction explicitly models interactions among anatomy-indexed tokens while retaining unrestricted global context. LGC therefore differs from mean pooling or direct summation of independently learned local features. To stabilize the composition, let $\mathcal B_v$ denote the valid samples containing at least one available organ. We introduce the following LGC consistency term:
\begin{equation}
\mathcal L_{\mathrm{lgc}}
=
1-\frac{1}{|\mathcal B_{\mathrm{v}}|}
\sum_{i\in\mathcal B_{\mathrm{v}}}
\cos\!\left(u_{g,i},a_{x,i}\right).
\label{eq:lgc}
\end{equation}

\subsection{Calibrate: Semantic Calibration of Evidence}

For compactness, we write symmetric batch-level contrast between representation batches $p$ and $q$ as
\begin{equation}
\begin{aligned}
\mathcal L_{\mathrm{ITC}}(p,q;\tau)
&=\tfrac12\!\left[
\mathrm{CE}(\bar p\bar q^\top/\tau,I)
+\mathrm{CE}(\bar q\bar p^\top/\tau,I)\right],\\[-1mm]
\bar p&=\mathrm{Norm}(p),\qquad
\bar q=\mathrm{Norm}(q).
\end{aligned}
\label{eq:symmetric_itc}
\end{equation}

Although $a_x$ has already been composed under global context, its semantics remain derived primarily from anatomy-specific evidence and are not automatically aligned with whole-study interpretation. We therefore use the intermediate report view to semantically calibrate $a_x$. This stage gives rise to the name \method: \emph{Semantic Calibration Of comPosed Evidence}.
The paired diagnostic summary is encoded as
$t_{\mathrm{imp}}=E_t(r_{\mathrm{imp}})$, and both modalities are projected into a shared embedding space:
\begin{equation}
\begin{aligned}
e_x&=g_e(a_x),\qquad
e_r=g_{\mathrm{imp}}(t_{\mathrm{imp}}),\\[-1mm]
\mathcal L_{\mathrm{evi}}
&=\mathcal L_{\mathrm{ITC}}(e_x,e_r;\tau_e).
\end{aligned}
\label{eq:evidence_itc}
\end{equation}
Here $g_e$ and $g_{\mathrm{imp}}$ are learned projection heads for visual evidence and diagnostic-summary text, respectively. Equation~\ref{eq:evidence_itc} directly supervises the composed visual evidence that will subsequently enter the residual pathway; it does not align the pre-LGC global query or use the text-side LGC representation as its target. The diagnostic summary is broader than an organ-specific snippet but more focused than the complete report, making it a natural supervisory target for bridging localized evidence and study-level interpretation.

\subsection{Integrate: Context-Preserving Residual Integration}

The semantically calibrated evidence is projected and incorporated into the direct whole-volume pathway:
\begin{equation}
e_{\mathrm{org}}=W_ea_x,\qquad
z_{\mathrm{SCOPE}}
=\mathrm{LN}\!\left(z_b+\gamma e_{\mathrm{org}}\right),
\label{eq:injection}
\end{equation}
where $W_e$ matches the embedding dimension and $\gamma$ controls the residual magnitude. Equation~\ref{eq:injection} preserves the unrestricted context encoded by $z_b$ while allowing composed and semantically calibrated organ evidence to modify the final representation. The contribution is therefore not generic global-local fusion or residual addition alone, but an explicit evidence transformation that makes anatomy-indexed information compatible with a whole-volume representation.

Within the residual model, $z_{\mathrm{SCOPE}}$ denotes the final residual-conditioned CT representation. Global image-report alignment is applied as:
\begin{equation}
\mathcal L_{\mathrm{global}}
=
\mathcal L_{\mathrm{ITC}}(z_{\mathrm{SCOPE}},z_r;\tau_g),
\label{eq:global_itc}
\end{equation}
where $\mathcal L_{\mathrm{ITC}}$ normalizes both modalities internally as defined in Eq.~\ref{eq:symmetric_itc}. The normalized representation
$\tilde z_{\mathrm{SCOPE}}=\mathrm{Norm}(z_{\mathrm{SCOPE}})$
is used as the final CT embedding for downstream evaluation.

\subsection{Auxiliary Semantic and Anatomical Objectives}

The organ-specific snippets yield text tokens
$T=\{E_t(r_o)\}_{o\in\mathcal O}$. Text-side LGC mirrors Eq.~\ref{eq:lgc}, using $t_{\mathrm{imp}}$ as the query and $T$ as the keys and values to produce an aggregated text representation $\bar t$. Text consistency and reconstruction encourage $\bar t$ to preserve the study-level semantics of $t_{\mathrm{imp}}$, providing auxiliary regularization across the report hierarchy. Separately, $u_g$ serves as decoder memory for teacher-forced diagnostic-summary reconstruction.

Anatomical supervision is provided through mask prediction and mask reconstruction from $a_x$. Together with text reconstruction and diagnostic-summary language modeling, these objectives provide complementary regularization for semantic coordination and anatomy-indexed representation learning. We use balanced binary cross-entropy for $\mathcal L_{\mathrm{mask}}$ and $\mathcal L_{\mathrm{mask\text{-}rec}}$, mean-squared error for $\mathcal L_{\mathrm{text\text{-}rec}}$, cosine distance for $\mathcal L_{\mathrm{text\text{-}lgc}}$, and token-level cross-entropy for $\mathcal L_{\mathrm{imp}}$.

\subsection{Training Objectives}

The image pathway first receives a short image-only supervised warm-up with disease classification, organ-concept prediction, mask prediction, mask reconstruction, and image-side LGC objectives. The anatomy supervision follows established 3D segmentation practice~\cite{hatamizadeh2022unetr,wasserthal2023totalsegmentator}; mask reconstruction and image-side LGC are model-specific warm-up objectives. Only the encoder and context-refinement pathway are transferred, while the warm-up query and LGC heads are discarded.

During the main alignment stage, the transferred encoder and context-refinement modules are frozen, while the newly initialized anatomy-indexed query transformer, image- and text-side LGC modules, evidence projection, residual integration, text encoder, and projection heads are optimized.
We group the training objectives by function: global image-report alignment, semantic calibration of composed evidence, organ-specific grounding, and auxiliary regularization. The complete objective is:
\begin{equation}
\begin{aligned}
\mathcal L={}&
\mathcal L_{\mathrm{global}}
+\lambda_{\mathrm{evi}}\mathcal L_{\mathrm{evi}}
+\lambda_{\mathrm{local}}\mathcal L_{\mathrm{local}}
+\lambda_{\mathrm{lgc}}\mathcal L_{\mathrm{lgc}}\\
&+\lambda_{\mathrm{mask}}\mathcal L_{\mathrm{mask}}
+\lambda_{\mathrm{rec}}\mathcal L_{\mathrm{mask\text{-}rec}}\\
&+\lambda_{\mathrm{tr}}\mathcal L_{\mathrm{text\text{-}rec}}
+\lambda_{\mathrm{tl}}\mathcal L_{\mathrm{text\text{-}lgc}}
+\lambda_{\mathrm{imp}}\mathcal L_{\mathrm{imp}}.
\end{aligned}
\label{eq:total_loss}
\end{equation}
Here $\lambda_{\mathrm{tr}}$ and $\lambda_{\mathrm{tl}}$ weight text reconstruction and text-LGC consistency, respectively. $\mathcal L_{\mathrm{global}}$, $\mathcal L_{\mathrm{evi}}$, and $\mathcal L_{\mathrm{local}}$ provide global, composed-evidence, and organ-level visual-text alignment. $\mathcal L_{\mathrm{lgc}}$ regularizes Local-Global evidence composition, while the remaining objectives provide anatomical grounding and reconstruction-based regularization.

\section{Experiments}
\label{sec:experiments}

\begin{table*}[!t]
\centering
\small
\setlength{\tabcolsep}{7.1pt}
\begin{tabular}{lcccccccc}
\toprule
Method & \multicolumn{4}{c}{CT-RATE} & \multicolumn{4}{c}{\radchest} \\
\cmidrule(lr){2-5}\cmidrule(l){6-9}
& AUROC & Accuracy & F1-score & Precision
& AUROC & Accuracy & F1-score & Precision \\
\midrule
CT-Net~\cite{draelos2021radchestct}
& 60.3 & 58.1 & 63.1 & 23.9
& 54.4 & 54.0 & 58.7 & 28.5 \\

CT-CLIP~\cite{hamamci2024generalist}
& 73.1 & 66.8 & 70.7 & 32.3
& 62.9 & 59.5 & 64.2 & 33.6 \\

BIUD~\cite{cao2024biud}
& 71.3 & 68.1 & 71.6 & 33.8
& 62.9 & 60.6 & 65.2 & 33.7 \\

Merlin~\cite{langlotz2024merlin}
& 72.8 & 67.2 & 70.9 & 33.7
& 64.4 & 61.9 & 66.3 & 34.8 \\

COLIPRI-C~\cite{wald2025colipri}
& 76.3 & - & - & -
& 69.1 & - & - & - \\

fVLM~\cite{shui2025fvlm}
& 77.8 & 71.8 & 75.1 & 37.9
& 68.0 & 64.7 & 68.8 & 37.4 \\
\midrule
\textbf{\method{} (Ours)}
& \textbf{85.0} & \textbf{78.9} & \textbf{81.0} & \textbf{45.7}
& \textbf{72.2} & \textbf{68.5} & \textbf{71.8} & \textbf{42.7} \\
\bottomrule
\end{tabular}
\caption{Zero-shot abnormality diagnosis on CT-RATE and \radchest. All metrics are reported as percentages (\%).}
\label{tab:diagnosis_results}
\end{table*}

\begin{table}[!t]
\centering
\small
\setlength{\tabcolsep}{2.1pt}
\begin{tabular}{lcccc}
\toprule
& \multicolumn{2}{c}{CT-RATE}
& \multicolumn{2}{c}{\radchest} \\
\cmidrule(lr){2-3}\cmidrule(l){4-5}
Method
& \small AUPRC & \small AUROC
& \small  AUPRC & \small AUROC \\
\midrule
Curia~\cite{dancette2025curia}
& 45.9 & 78.0 & 39.2 & 65.7 \\

CT-CLIP
& - & 75.1 & - & 64.7 \\

CT-FM~\cite{ct-fm}
& 53.5 & 82.1 & 42.4 & 68.5 \\

Merlin
& 54.8 & 82.6 & 45.3 & 70.9 \\

COLIPRI-C
& 55.1 & 82.6 & 46.3 & 71.2 \\
\midrule
\textbf{\method{} (Ours)}
& \textbf{64.5} & \textbf{87.0}
& \textbf{55.2} & \textbf{76.6} \\
\bottomrule
\end{tabular}
\caption{Frozen-encoder linear probing on CT-RATE and \radchest. AUPRC and AUROC are reported as percentages (\%).}
\label{tab:linear_probe}
\end{table}

\subsection{Experimental Setup}

\paragraph{Datasets.} \ctrate contains 25,692 non-contrast chest CT scans from 21,304 patients, expanded to 50,188 reconstructed volumes with paired reports and labels for 18 abnormalities~\cite{hamamci2024generalist}. We use 24,128 processed training records with CT volumes, organ masks, and structured organ-report fields. \radchest is used for external evaluation with 16 abnormalities categories on 3,630 public scans~\cite{draelos2021radchestct}.

\paragraph{Evaluation.}
We evaluate zero-shot abnormality diagnosis, image-to-image retrieval, report-to-image retrieval, and frozen-encoder linear probing. Diagnosis is assessed using AUC, accuracy, F1 score, and precision; retrieval uses mAP@5/10/50 and Recall@5/10/50/100. Unless otherwise specified, metrics are reported as percentages, while organ-occlusion sensitivity results are presented as qualitative comparisons. Missing baseline results are denoted by ``-''.

\paragraph{Implementation.}
CT volumes are intensity-normalized and resized by cropping or padding to $96\times192\times192$. We use a 3D ResNet-18~\cite{hara2018r3d} as the image encoder and BiomedVLP CXR-BERT~\cite{boecking2022cxrbert} as the text encoder. Models are optimized with AdamW using a learning rate of $10^{-5}$, gradient clipping of $0.5$, and a batch size of 4. The loss weights for global ITC, evidence ITC, local ITC, image-side LGC, mask grounding, mask reconstruction, text reconstruction, text-LGC consistency, and diagnostic-summary language modeling are $1.0$, $0.02$, $0.2$, $0.5$, $1.0$, $1.0$, $0.2$, $0.3$, and $0.1$, respectively. Zero-shot diagnosis uses normalized image-text similarity without training a downstream classifier. All experiments were conducted on NVIDIA A40 GPUs.

\subsection{Main Results}

\paragraph{Abnormality diagnosis.}
As shown in Table~\ref{tab:diagnosis_results}, \method achieves an AUC of 85.0\%, an accuracy of 78.9\%, an F1 score of 81.0\%, and a precision of 45.7\% on CT-RATE. On \radchest, the corresponding results are 72.2\%, 68.5\%, 71.8\%, and 42.7\%. On CT-RATE, \method improves over the global CT-CLIP baseline by 11.9\% AUC and the anatomy-level fVLM baseline by 7.2\%; the corresponding gains on \radchest are 9.3\% and 4.2\%. These improvements over both global and anatomy-level baselines support the benefit of composing anatomy-indexed evidence under unrestricted whole-volume context rather than relying on either representation level alone.

Table~\ref{tab:linear_probe} further evaluates representation transfer through frozen-encoder linear probing. \method achieves 87.0\% AUROC on CT-RATE and 76.6\% on \radchest, indicating that the learned representation preserves transferable diagnostic information through complementary whole-volume context and anatomy-indexed evidence.
Together, the diagnostic and linear-probe results show that the proposed representation is effective for both downstream prediction and frozen-feature transfer.

\paragraph{Image and report retrieval.}

\begin{table}[!t]
\small
\centering
\setlength{\tabcolsep}{4pt}
\begin{tabular}{lll}
\toprule
Method 
& \shortstack{Image$\rightarrow$Image \\ mAP@5/10/50 (\%)}
& \shortstack{Report$\rightarrow$Image \\ R@5/10/50/100 (\%)} \\
\midrule
CT-Net
& 59.4 / 48.1 / 40.7
& - \\

Merlin
& 62.6 / 51.3 / 43.9
& 1.5 / 2.7 / 7.7 / 12.7 \\

CT-CLIP
& 68.3 / 57.2 / 48.9
& 2.9 / 5.0 / 18.0 / 28.7 \\

fVLM
& 49.1 / 36.8 / 26.0
& 0.8 / 1.5 / 4.9 / 8.3 \\
\midrule

\textbf{\method{} (Ours)}
& \textbf{70.8 / 61.1 / 53.6}
& \textbf{10.8 / 17.6 / 43.0 / 56.6} \\
\bottomrule
\end{tabular}
\caption{Image-to-image and report-to-image retrieval on CT-RATE. All metrics are reported as percentages.}
\label{tab:retrieval_results}
\end{table}

Table~\ref{tab:retrieval_results} evaluates whether the residual-conditioned representation preserves clinically meaningful image neighborhoods and cross-modal correspondence. For image-to-image retrieval, \method obtains mAP values of 70.8\%, 61.1\%, and 53.6\% at $K=5$, $10$, and $50$, improving over CT-CLIP by 2.5, 3.9, and 4.7\%, respectively. For report-to-image retrieval, \method achieves Recall values of 10.8\%, 17.6\%, 43.0\%, and 56.6\% at $K=5$, $10$, $50$, and $100$, improving over CT-CLIP by 7.9, 12.6, 25.0, and 27.9\%. These gains indicate that the learned representation preserves whole-examination structure while maintaining strong image-report alignment.

\subsection{Ablation and Analysis}
\paragraph{Evidence composition and calibration.}
Table~\ref{tab:ablation} evaluates the proposed \emph{Ground-Compose-Calibrate-Integrate} pathway through cumulative stage-wise ablations. \emph{Grounding} combines organ-level text contrast with mask-based anatomical supervision, while \emph{Composition} combines image-side local-global composition with text-side aggregation and reconstruction. \emph{Calibration} aligns the composed evidence with the diagnostic summary, and \emph{Integration} fuses the calibrated evidence with the whole-volume representation.
The whole-volume baseline achieves 78.3\% AUC. Adding grounding, composition, calibration, and integration progressively improves AUC to 81.2\%, 83.7\%, 84.1\%, and 85.0\%, respectively. Overall, the complete pathway yields a 6.7-percentage-point improvement over the whole-volume baseline, demonstrating the complementary contribution of the four stages.

\begin{table}[!t]
\centering
\small
\setlength{\tabcolsep}{2.5pt}
\begin{tabular}{@{}lccccc@{}}
\toprule
Configuration
& Ground.
& Compose
& Calib.
& Integr.
& AUC\\
\midrule
Whole-volume
& - & - & - & - & 78.3 \\

+ Anatomy grounding
& \checkmark & - & - & - & 81.2 \\

+ Evidence composition
& \checkmark & \checkmark & - & - & 83.7 \\

+ Semantic calibration
& \checkmark & \checkmark & \checkmark & - & 84.1 \\

+ Residual integration
& \checkmark & \checkmark & \checkmark & \checkmark & \textbf{85.0} \\
\bottomrule
\end{tabular}

\caption{\textbf{Stage-wise ablation of the Ground-Compose-Calibrate-Integrate pathway on CT-RATE.}
\textit{Ground.} combines organ-level text contrast with mask-based anatomical supervision;
\textit{Compose} combines image-side local-global composition with text-side aggregation and reconstruction;
\textit{Calib.} denotes semantic calibration of the composed evidence using the diagnostic summary; and
\textit{Integr.} denotes residual integration with the whole-volume representation.
Each row cumulatively includes all preceding stages.}
\label{tab:ablation}
\end{table}

\paragraph{Complementary global and anatomy-indexed evidence.}
\begin{table}[!t]
\centering
\small
\setlength{\tabcolsep}{4.6pt}
\renewcommand{\arraystretch}{1.05}

\begin{tabular}{@{}lccc@{\hspace{8pt}}ccc@{}}
\toprule
& \multicolumn{3}{c}{\textbf{Integrated rep.}}
& \multicolumn{3}{c}{\textbf{Organ-query rep.}} \\
\cmidrule(lr){2-4}\cmidrule(lr){5-7}

\textbf{Group}
& \method & Vol. & $\Delta$
& \method & fVLM & $\Delta$ \\
\midrule

Macro AUC
& \textbf{85.0} & 78.3 & +6.7
& \textbf{79.9} & 77.9 & +2.0 \\

\midrule

Lung
& 82.3 & 75.4 & +6.9
& 75.2 & 74.3 & +0.9 \\

Heart
& 93.1 & 89.3 & +3.8
& 89.9 & 88.3 & +1.6 \\

Esophagus
& 85.8 & 70.2 & +15.6
& 86.3 & 74.3 & +12.0 \\

Aorta
& 94.2 & 91.1 & +3.1
& 94.4 & 89.2 & +5.2 \\

Other
& 83.1 & 75.1 & +8.0
& - & - & - \\

\bottomrule
\end{tabular}

\caption{\textbf{Does anatomy-indexed evidence improve both integrated and organ-specific representations?}
The \textit{integrated representation} block compares the complete \method with the whole-volume variant (Vol.), which removes the anatomy-indexed evidence pathway, to test whether organ-level evidence provides complementary information beyond global volume features.
The \textit{organ-query representation} block compares the learned organ queries with our fVLM reproduction on 16 organ-associated concepts to test whether they encode stronger organ-specific disease semantics.}
\label{tab:anatomy_evidence}
\end{table}

Table~\ref{tab:anatomy_evidence} evaluates whether anatomy-indexed evidence improves the integrated representation and whether the learned organ queries encode stronger organ-specific semantics.

The left part compares \method with a whole-volume variant without the anatomy-indexed pathway. \method improves macro AUC from 78.3\% to 85.0\%, with gains across all anatomical groups, including an 8.0-point improvement on the ``Other'' group, suggesting benefits beyond the explicitly queried organs. The right part evaluates the organ-query representations against fVLM on 16 organ-associated concepts. \method improves macro AUC from 77.9\% to 79.9\% and consistently outperforms fVLM across all four organs. Together, these results show that \method learns stronger anatomy-specific representations and effectively integrates them with global evidence.

\begin{figure}[!t]
\centering
\includegraphics[width=\columnwidth]{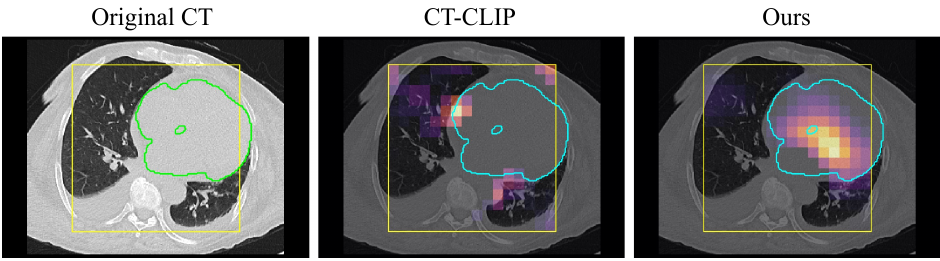}
\caption{Organ-occlusion sensitivity for a heart-associated finding. \method concentrates disease-score changes within the target cardiac region.}
\label{fig:occlusion_sensitivity}
\end{figure}

\paragraph{Organ-occlusion sensitivity.}
To qualitatively assess whether the learned representation attends to anatomically relevant evidence, we conduct an organ-occlusion sensitivity analysis for a heart-associated finding. Specifically, local regions of the input CT are sequentially occluded, and the resulting absolute changes in the predicted disease score are visualized on a representative axial slice. As shown in Figure~\ref{fig:occlusion_sensitivity}, CT-CLIP yields diffuse responses, including sensitivity beyond the target cardiac region. By contrast, \method produces a more compact response concentrated within the heart, with reduced sensitivity to irrelevant surrounding tissues. These results suggest that grounding, composition, semantic calibration, and integration jointly promote anatomically consistent associations between disease predictions and supporting evidence, qualitatively complementing the quantitative performance gains observed across benchmarks.

\section{Conclusion}

We presented \method, a framework for \emph{Semantic Calibration Of comPosed Evidence} in CT vision-language learning. Through \emph{Ground-Compose-Calibrate-Integrate}, \method composes anatomy-indexed evidence under whole-volume context, calibrates it toward study-level semantics, and integrates it into a context-preserving global representation. Across diverse evaluation tasks, \method consistently improves upon prior methods: it achieves 85.0\% and 72.2\% AUC for zero-shot diagnosis on CT-RATE and \radchest, improving over fVLM by 7.2 and 4.2 points, respectively, and reaches 87.0\% and 76.6\% AUROC in frozen-encoder linear probing. For retrieval, \method surpasses the strongest prior results by up to 4.7 mAP points for image-to-image retrieval and 27.9 Recall points for report-to-image retrieval. Organ-level discrimination and occlusion analyses further demonstrate stronger anatomy-indexed representations and selective use of localized evidence. Together, these results establish semantic calibration of composed evidence as an effective strategy for transferable CT vision-language representation learning.

\bibliography{references}

@article{hamamci2024generalist,
  title={Generalist foundation models from a multimodal dataset for 3D computed tomography},
  author={Hamamci, Ibrahim Ethem and Er, Sezgin and Wang, Chenyu and Almas, Furkan and Simsek, Ayse Gulnihan and Esirgun, Sevval Nil and Dogan, Irem and Durugol, Omer Faruk and Hou, Benjamin and Shit, Suprosanna and others},
  journal={Nature Biomedical Engineering},
  pages={1--19},
  year={2026},
  publisher={Nature Publishing Group UK London}
}

@inproceedings{boecking2022cxrbert,
  title={Making the most of text semantics to improve biomedical vision--language processing},
  author={Boecking, Benedikt and Usuyama, Naoto and Bannur, Shruthi and Castro, Daniel C and Schwaighofer, Anton and Hyland, Stephanie and Wetscherek, Maria and Naumann, Tristan and Nori, Aditya and Alvarez-Valle, Javier and others},
  booktitle={European conference on computer vision},
  pages={1--21},
  year={2022},
  organization={Springer}
}

@inproceedings{shui2025fvlm,
  title={Large-scale and Fine-grained Vision-language Pre-training for Enhanced CT Image Understanding},
  author={Shui, Zhongyi and Zhang, Jianpeng and Cao, Weiwei and Wang, Sinuo and Guo, Ruizhe and Lu, Le and Yang, Lin and Ye, Xianghua and Liang, Tingbo and Zhang, Qi and others},
  booktitle={The Thirteenth International Conference on Learning Representations},
  year =          {2025},
}

@inproceedings{liu2023t3d,
  title={T3D: Advancing 3D Medical Vision-Language Pre-training by Learning Multi-View Visual Consistency},
  author={Liu, Che and Ouyang, Cheng and Chen, Yinda and Quilodr{\'a}n-Casas, C{\'e}sar and Ma, Lei and Fu, Jie and Guo, Yike and Shah, Anand and Bai, Wenjia and Arcucci, Rossella},
  booktitle={Proceedings of the IEEE/CVF International Conference on Computer Vision},
  pages={6704--6714},
  year={2025}
}

@article{langlotz2024merlin,
  title={Merlin: a computed tomography vision--language foundation model and dataset},
  author={Blankemeier, Louis and Kumar, Ashwin and Cohen, Joseph Paul and Liu, Jiaming and Liu, Longchao and Van Veen, Dave and Gardezi, Syed Jamal Safdar and Yu, Hongkun and Paschali, Magdalini and Chen, Zhihong and others},
  journal={Nature},
  volume={652},
  number={8112},
  pages={1318--1328},
  year={2026},
  publisher={Nature Publishing Group UK London}
}

@article{liu2025visual,
  title={Visual--language foundation models in medicine},
  author={Liu, Chunyu and Jin, Yixiao and Guan, Zhouyu and Li, Tingyao and Qin, Yiming and Qian, Bo and Jiang, Zehua and Wu, Yilan and Wang, Xiangning and Zheng, Ying Feng and others},
  journal={The Visual Computer},
  volume={41},
  number={4},
  pages={2953--2972},
  year={2025},
  publisher={Springer}
}

@article{wald2025colipri,
  title={Comprehensive language-image pre-training for 3D medical image understanding},
  author={Wald, Tassilo and Hamamci, Ibrahim Ethem and Gao, Yuan and Bond-Taylor, Sam and Sharma, Harshita and Ilse, Maximilian and Lo, Cynthia and Melnichenko, Olesya and Schwaighofer, Anton and Codella, Noel CF and others},
  journal={arXiv preprint arXiv:2510.15042},
  year={2025}
}

@inproceedings{huang2021gloria,
  title={Gloria: A multimodal global-local representation learning framework for label-efficient medical image recognition},
  author={Huang, Shih-Cheng and Shen, Liyue and Lungren, Matthew P and Yeung, Serena},
  booktitle={Proceedings of the IEEE/CVF international conference on computer vision},
  pages={3942--3951},
  year={2021}
}

@inproceedings{muller2022joint,
  title={Joint learning of localized representations from medical images and reports},
  author={M{\"u}ller, Philip and Kaissis, Georgios and Zou, Congyu and Rueckert, Daniel},
  booktitle={European conference on computer vision},
  pages={685--701},
  year={2022},
  organization={Springer}
}

@inproceedings{zhang2022convirt,
  title={Contrastive learning of medical visual representations from paired images and text},
  author={Zhang, Yuhao and Jiang, Hang and Miura, Yasuhide and Manning, Christopher D and Langlotz, Curtis P},
  booktitle={Machine learning for healthcare conference},
  pages={2--25},
  year={2022},
  organization={PMLR}
}

@article{tiu2022chexzero,
  title={Expert-level detection of pathologies from unannotated chest X-ray images via self-supervised learning},
  author={Tiu, Ekin and Talius, Ellie and Patel, Pujan and Langlotz, Curtis P and Ng, Andrew Y and Rajpurkar, Pranav},
  journal={Nature biomedical engineering},
  volume={6},
  number={12},
  pages={1399--1406},
  year={2022},
  publisher={Nature Publishing Group UK London}
}

@inproceedings{wang2022medclip,
  title={Medclip: Contrastive learning from unpaired medical images and text},
  author={Wang, Zifeng and Wu, Zhenbang and Agarwal, Dinesh and Sun, Jimeng},
  booktitle={Proceedings of the 2022 Conference on Empirical Methods in Natural Language Processing},
  pages={3876--3887},
  year={2022}
}

@article{zhou2022refers,
  title={Generalized radiograph representation learning via cross-supervision between images and free-text radiology reports},
  author={Zhou, Hong-Yu and Chen, Xiaoyu and Zhang, Yinghao and Luo, Ruibang and Wang, Liansheng and Yu, Yizhou},
  journal={Nature Machine Intelligence},
  volume={4},
  number={1},
  pages={32--40},
  year={2022},
  publisher={Nature Publishing Group UK London}
}

@inproceedings{chen2022arl,
  title={Align, reason and learn: Enhancing medical vision-and-language pre-training with knowledge},
  author={Chen, Zhihong and Li, Guanbin and Wan, Xiang},
  booktitle={Proceedings of the 30th ACM international conference on multimedia},
  pages={5152--5161},
  year={2022}
}

@inproceedings{chen2022m3ae,
  title={Multi-modal masked autoencoders for medical vision-and-language pre-training},
  author={Chen, Zhihong and Du, Yuhao and Hu, Jinpeng and Liu, Yang and Li, Guanbin and Wan, Xiang and Chang, Tsung-Hui},
  booktitle={International Conference on Medical Image Computing and Computer-Assisted Intervention},
  pages={679--689},
  year={2022},
  organization={Springer}
}

@inproceedings{wu2023medklip,
  title={Medklip: Medical knowledge enhanced language-image pre-training for x-ray diagnosis},
  author={Wu, Chaoyi and Zhang, Xiaoman and Zhang, Ya and Wang, Yanfeng and Xie, Weidi},
  booktitle={Proceedings of the IEEE/CVF international conference on computer vision},
  pages={21372--21383},
  year={2023}
}

@inproceedings{chen2023ptunifier,
  title={Towards unifying medical vision-and-language pre-training via soft prompts},
  author={Chen, Zhihong and Diao, Shizhe and Wang, Benyou and Li, Guanbin and Wan, Xiang},
  booktitle={Proceedings of the IEEE/CVF international conference on computer vision},
  pages={23403--23413},
  year={2023}
}

@inproceedings{hara2018r3d,
  title={Can spatiotemporal 3d cnns retrace the history of 2d cnns and imagenet?},
  author={Hara, Kensho and Kataoka, Hirokatsu and Satoh, Yutaka},
  booktitle={Proceedings of the IEEE conference on Computer Vision and Pattern Recognition},
  pages={6546--6555},
  year={2018}
}

@inproceedings{hatamizadeh2022unetr,
  title={Unetr: Transformers for 3d medical image segmentation},
  author={Hatamizadeh, Ali and Tang, Yucheng and Nath, Vishwesh and Yang, Dong and Myronenko, Andriy and Landman, Bennett and Roth, Holger R and Xu, Daguang},
  booktitle={Proceedings of the IEEE/CVF winter conference on applications of computer vision},
  pages={574--584},
  year={2022}
}

@article{wasserthal2023totalsegmentator,
  title={TotalSegmentator: robust segmentation of 104 anatomic structures in CT images},
  author={Wasserthal, Jakob and Breit, Hanns-Christian and Meyer, Manfred T and Pradella, Maurice and Hinck, Daniel and Sauter, Alexander W and Heye, Tobias and Boll, Daniel T and Cyriac, Joshy and Yang, Shan and others},
  journal={Radiology: Artificial Intelligence},
  volume={5},
  number={5},
  pages={e230024},
  year={2023},
  publisher={Radiological Society of North America}
}

@article{draelos2021radchestct,
  title={Machine-learning-based multiple abnormality prediction with large-scale chest computed tomography volumes},
  author={Draelos, Rachel Lea and Dov, David and Mazurowski, Maciej A and Lo, Joseph Y and Henao, Ricardo and Rubin, Geoffrey D and Carin, Lawrence},
  journal={Medical image analysis},
  volume={67},
  pages={101857},
  year={2021},
  publisher={Elsevier}
}

@inproceedings{cao2024biud,
  title={Bootstrapping chest ct image understanding by distilling knowledge from x-ray expert models},
  author={Cao, Weiwei and Zhang, Jianpeng and Xia, Yingda and Mok, Tony CW and Li, Zi and Ye, Xianghua and Lu, Le and Zheng, Jian and Tang, Yuxing and Zhang, Ling},
  booktitle={Proceedings of the IEEE/CVF Conference on Computer Vision and Pattern Recognition},
  pages={11238--11247},
  year={2024}
}

@inproceedings{chen2023cancerunit,
  title={CancerUniT: Towards a Single Unified Model for Effective Detection, Segmentation, and Diagnosis of Eight Major Cancers Using a Large Collection of CT Scans},
  author={Chen, Jieneng and Xia, Yingda and Yao, Jiawen and Yan, Ke and Zhang, Jianpeng and Lu, Le and Wang, Fakai and Zhou, Bo and Qiu, Mingyan and Yu, Qihang and Yuan, Mingze and Fang, Wei and Tang, Yuxing and Xu, Minfeng and Zhou, Jian and Zhao, Yuqian and Wang, Qifeng and Ye, Xianghua and Yin, Xiaoli and Shi, Yu and Chen, Xin and Zhou, Jingren and Yuille, Alan and Liu, Zaiyi and Zhang, Ling},
  booktitle={Proceedings of the IEEE/CVF International Conference on Computer Vision},
  pages={21327--21338},
  year={2023}
}

@article{zhang2025multimodal,
  title={A multimodal biomedical foundation model trained from fifteen million image--text pairs},
  author={Zhang, Sheng and Xu, Yanbo and Usuyama, Naoto and Xu, Hanwen and Bagga, Jaspreet and Tinn, Robert and Preston, Sam and Rao, Rajesh and Wei, Mu and Valluri, Naveen and others},
  journal={Nejm Ai},
  volume={2},
  number={1},
  pages={AIoa2400640},
  year={2025},
  publisher={Massachusetts Medical Society}
}

@article{ct-fm,
  title={Vision foundation models for computed tomography},
  author={Pai, Suraj and Hadzic, Ibrahim and Bontempi, Dennis and Bressem, Keno and Kann, Benjamin H and Fedorov, Andriy and Mak, Raymond H and Aerts, Hugo JWL},
  journal={arXiv preprint arXiv:2501.09001},
  year={2025}
}

@article{codella2024medimageinsight,
  title={Medimageinsight: An open-source embedding model for general domain medical imaging},
  author={Codella, Noel CF and Jin, Ying and Jain, Shrey and Gu, Yu and Lee, Ho Hin and Abacha, Asma Ben and Santamaria-Pang, Alberto and Guyman, Will and Sangani, Naiteek and Zhang, Sheng and others},
  journal={arXiv preprint arXiv:2410.06542},
  year={2024}
}

@article{dancette2025curia,
  title={Curia: A Multi-Modal Foundation Model for Radiology},
  author={Dancette, Corentin and Khlaut, Julien and Saporta, Antoine and Philippe, Helene and Ferreres, Elodie and Callard, Baptiste and Danielou, Th{\'e}o and Alberge, L{\'e}o and Machado, L{\'e}o and Tordjman, Daniel and others},
  journal={arXiv preprint arXiv:2509.06830},
  year={2025}
}

@article{wang2026costal,
  title={Costal Cartilage Segmentation with Topology Guided Deformable Mamba: Method and Benchmark},
  author={Wang, Senmao and Gong, Haifan and Cui, Runmeng and Wan, Boyao and Hu, Zhonglin and Yang, Haiqing and Zhou, Jingyang and Jiang, Haiyue and Lin, Lin},
  journal={Expert Systems with Applications},
  volume={300},
  pages={130085},
  year={2026}
}

@article{gong2025boundary,
  title={Boundary as the Bridge: Toward Heterogeneous Partially-Labeled Medical Image Segmentation and Landmark Detection},
  author={Gong, Haifan and Wan, Boyao and Kang, Luoyao and Wan, Xiang and Zhang, Lingyan and Li, Haofeng},
  journal={IEEE Transactions on Medical Imaging},
  volume={44},
  number={7},
  pages={2747--2756},
  year={2025}
}

@article{huang2025bcnet,
  title={BCNet: Bronchus Classification via Structure Guided Representation Learning},
  author={Huang, Wenhao and Gong, Haifan and Zhang, Huan and Wang, Yu and Wan, Xiang and Li, Guanbin and Li, Haofeng and Shen, Hong},
  journal={IEEE Transactions on Medical Imaging},
  volume={44},
  number={1},
  pages={489--498},
  year={2025}
}

@article{gong2022vqamix,
  title={VQAMix: Conditional Triplet Mixup for Medical Visual Question Answering},
  author={Gong, Haifan and Chen, Guanqi and Mao, Mingzhi and Li, Zhen and Li, Guanbin},
  journal={IEEE Transactions on Medical Imaging},
  volume={41},
  number={11},
  pages={3332--3343},
  year={2022}
}

@inproceedings{gong2025domain,
  title={Domain Generalized Medical Landmark Detection via Robust Boundary-Aware Pre-Training},
  author={Gong, Haifan and Lu, Yu and Wan, Xiang and Li, Haofeng},
  booktitle={Proceedings of the AAAI Conference on Artificial Intelligence},
  volume={39},
  number={3},
  pages={3140--3148},
  year={2025}
}

@inproceedings{gong2024intensity,
  title={Intensity Confusion Matters: An Intensity-Distance Guided Loss for Bronchus Segmentation},
  author={Gong, Haifan and Huang, Wenhao and Zhang, Huan and Wang, Yu and Wan, Xiang and Shen, Hong and Li, Guanbin and Li, Haofeng},
  booktitle={2024 IEEE International Conference on Multimedia and Expo (ICME)},
  pages={1--6},
  year={2024}
}

@inproceedings{gong2021crossmodal,
  title={Cross-Modal Self-Attention with Multi-Task Pre-Training for Medical Visual Question Answering},
  author={Gong, Haifan and Chen, Guanqi and Liu, Sishuo and Yu, Yizhou and Li, Guanbin},
  booktitle={Proceedings of the 2021 International Conference on Multimedia Retrieval},
  pages={456--460},
  year={2021},
  publisher={ACM}
}

@article{dong2022survey,
  title={A Survey of Natural Language Generation},
  author={Dong, Chenhe and Li, Yinghui and Gong, Haifan and Chen, Miaoxin and Li, Junxin and Shen, Ying and Yang, Min},
  journal={ACM Computing Surveys},
  volume={55},
  number={8},
  articleno={173},
  numpages={38},
  year={2022}
}

\end{document}